\documentclass[conference]{IEEEtran}

\usepackage{graphicx,times,psfig,amsmath} 
\usepackage{epsfig}

\begin{document}

\newtheorem{definition}{Definition}[section]
\newtheorem{lemma}{Lemma}[section]
\newtheorem{theorem}{Theorem}[section]
\newtheorem{property}{Property}[section]

\title{A New Kind of Hopfield Networks \\ for Finding Global Optimum}

\author{\authorblockN{Xiaofei Huang}
\authorblockA{School of Information Science and Technology\\
Tsinghua University, Beijing, P.~R.~China, 100084 \\
huangxiaofei@ieee.org}}
%


%


\maketitle

\begin{abstract}
The Hopfield network has been applied to solve optimization problems over decades.
However, it still has many limitations in accomplishing this task.
Most of them are inherited from the optimization algorithms it implements.
The computation of a Hopfield network, defined by a set of difference equations,
   can easily be trapped into one local optimum or another,
   sensitive to initial conditions, perturbations, and neuron update orders.
It doesn't know how long it will take to converge, as well as if the final solution is a global optimum, or not.
In this paper, we present a Hopfield network with a new set of difference equations
   to fix those problems.
The difference equations directly implement a new powerful optimization algorithm.
\footnote{Accepted by International Joint Conference on Neural Networks, 2005}
\end{abstract}


\section{Introduction}
In the beginning of the 1980s, Hopfield~\cite{Hopfield82,Hopfield85} and his colleges published two scientific papers 
   on ``neuron" computation.
Hopfield showed that highly interconnected networks of nonlinear analog neurons 
   are extremely effective in solving optimization problems.
From that time on,
   people has being applying the Hopfield network to solve a wide class of 
   combinatorial optimization problems (see a survey~\cite{Smith99}).

In a discrete-time version, the Hopfield network implemented 
   local search.
In a continuous-time version, it implemented gradient decent.
Both algorithms suffer the local minimum problem 
   and many optimization problems in practice have lots of local minima.
Furthermore, the Hopfield-Tank formulation of the energy function of the network
   causes infeasible solutions to occur most of the time~\cite{Wilson88,Smith99}.
People also found that those valid solutions were only slightly better than randomly chosen ones.

To guarantee the feasibility of the solutions,
   the most important breakthrough came from the valid subspace approaches of Aiyer et al~\cite{Aiyer90} 
   and Gee~\cite{Gee95}.
However, it requires researchers to design a constraint energy function to make solution feasible,
   add it to the original energy function,
   and recalculate the energy function to obtain new connection weights.
It is not simple and is unlikely that biological neural networks also implement such a process.
To escape from local minima,
   many variations of the Hopfield network 
   have been proposed based on the principles of simulated annealing~\cite{Kirkpatrick83}.
Three major approaches are Boltzmann~\cite{Hinton84}, Cauchy~\cite{Jeong89}, and Gaussian Machines~\cite{Akiyama89}.
In theory, simulated annealing can approach the global optimal solution
   in exponential time.
However, it is not guaranteed and is very slow to make it effective in practice.
Like local search, 
   it doesn't know how long it will take to converge.
It also does not know if a solution is a global optimum so that the search process can be stopped.

Those improvements make the Hopfield network competitive with conventional optimization algorithms,
   such as simulated annealing.
However, it can not be more powerful than those algorithms because it is just the implementations
   of those algorithms using interconnected networks of computing units, such as neurons.
Its capability is restricted due to 
   the limitations of the network structure and the theoretical limitations of
   the optimization algorithms it implements.
Those conventional optimization algorithms have both performance problems and convergence problems, 
   far from satisfactory in solving problems in practice.
For example, stereo matching is an important problem in computer vision,
   and one of the most active research areas in that field~\cite{Scharstein2002,Boykov2001,Lawrence2000,Marr76}.
Compared with many specialized optimization algorithms, such as graph cuts~\cite{Boykov2001,Kolmogorov04},
   simulated annealing has the worst performance in both the solution quality and computing time.
People in the computer vision community even do not need to put it 
   in the comparison list in the evaluation of different optimization algorithms for stereo matching~\cite{Scharstein2002}.   
   
In this paper, we present a Hopfield network with a new set of difference equations
   to fix those problems.
In solving large scale optimization problems in computer vision, 
   it significantly outperform general optimization algorithms, such as simulated annealing 
   and local search with multi-restarts, as well as specialized algorithms.

\section{Cooperative Neural Network Computation}


One of the most popular energy functions used in computer vision and other areas has
   the following general form:
\begin{equation}
E(x_1,x_2,\ldots,x_n)=\sum_{i} C_i (x_{i}) + \sum_{i,j, i \not = j} C_{ij} (x_i, x_j)\ ,
\label{binary_cost_function}
\end{equation}
where each variable $x_i$, for $1 \le i \le n$,
   has a finite domain $D_i$ of size $m_i$ ($m_i=|D_i|$).
$C_i$ is a real-valued function on variable $x_i$, called the unary constraint on $x_i$ 
   and $C_{ij}$ is a real-valued function on variable $x_i$ and $x_j$, 
   called the binary constraint on $x_i$ and $x_j$.
The optimization of the above energy function is NP-hard.
It is called the binary constraint optimization problem 
   in computer science,
   a special case of constraint model 
   where each constraint involves at most two variables.

Without loss of generality,
   we assume all constraints are nonnegative functions 
   through out this paper.
Also, we focus on the minimization of the function~(\ref{binary_cost_function})
   because minimization and maximization are logically equivalent.

\subsection{The New Set of Difference Equations}

Following the Hopfield network formulation, 
   we use a number of neurons,
   one for each value of each variable.
If variable $x_i$ has $m_i$ values, we have $m_i$ neurons, one for each value.
This set of neurons is called a neuron group.
In total, we have $\sum_i m_i$ neurons organized as $n$ groups for $n$ variables.

The state of the neuron is denoted as $c_i(x_i)$ for the value $x_i$ of the $i$th variable.
Because we are dealing with the minimization problem,
   for the convenience of the mathematical manipulation,
   we make $c_i(x_i) \ge 0$ and use lower values to indicate higher activation.
If $c_i(x_i) = 0$, the corresponding neuron has the highest activation level.

Different from the Hopfield network, 
   we use a new set of difference equations shown as follows:
\begin{equation}
c^{(k)}_i (x_i)   = \sigma_{t^{(k)}_i} g_i (x_i), \quad \mbox{for $x_i \in D_i$ and $1 \le i \le n$,}
\label{cooperative_optimization_binary}
\end{equation}
where $\sigma_{t^{(k)}_i}$ is a threshold function
   with the threshold $t^{(k)}_i$.
The threshold function $\sigma_t$ is defined as
\begin{displaymath}
\sigma_t(x) = \left \{\begin{array}{l}
        x,~~~~~\mbox{if $x \le t$;}\\
        \infty,~~~~\mbox{if $x > t$.}
       \end{array}
\right.
\end{displaymath}
and $g_i (x_i)$ is defined as 
\begin{eqnarray*}
\lefteqn{g_i (x_i)  = (1 - \lambda_k) C_i(x_i) + \lambda_k w_{ii} c^{(k-1)}_i (x_i) + } \\
   & & \sum_{j, j\not=i} \min_{x_j}\left((1 - \lambda_k ) C_{ij}(x_i, x_j) + \lambda_k w_{ij} c^{(k-1)}_j(x_j)\right), 
\end{eqnarray*}
where $k$ is the iteration step, 
   $w_{ij}$ are non-negative weights satisfying $\sum_i w_{ij} = 1$. 
Parameter $\lambda_k$ is a weight satisfying $0 \le \lambda_k < 1.$   

To ensure the feasibility of solutions, 
   we follow the winner-take-all principle by using the threshold function
   to progressively inhibit more and more neurons in the same group.
Eventually, we let only one neuron be active in one group.
Those values that the remaining active neurons corresponding to constitute a solution 
   to the original combinatorial optimization problem~(\ref{binary_cost_function}).
This approach is different from the one used by the Hopfield network.
The Hopfield network adds a new constraint energy function to ensure the feasibility of solutions.
   
This set of difference equations is a direct neural network implementation of 
   a new optimization algorithm (detail in the following subsection)
   for energy function minimization in a general form.
This algorithm is based a principle for optimization, called cooperative computation, 
   completely different from many existing ones.
Given an optimization problem instance, the computation always has a unique equilibrium 
   and converges to it with an exponential rate regardless of initial conditions and perturbations.

It is important to note that, different from the Hopfield network in a discrete-time version,
   our set of difference equations does not require to update one state at one time.
All states for all values of all variables can be updated simultaneously in parallel.
It is another advantage of this neural network over the classical Hopfield network.

\subsection{The Cooperative Optimization Algorithm}

To solve a hard combinatorial optimization problem, 
   we follow the divide-and-conquer principle.
We first break up the problem into a number of sub-problems of
   manageable sizes and complexities.
Following that, we assign each sub-problem to an agent,
	and ask those agents to solve the sub-problems in a cooperative way.
The cooperation is achieved by asking each agent
   to compromise its solution with the solutions of others
   instead of solving the sub-problems independently.
We can make an analogy with team playing, 
   where the team members work together
   to achieve the best for the team, but not necessarily the best for each member.
In many cases, cooperation of this kind can dramatically improve the problem-solving capabilities
	of the agents as a team, even when each agent may have very limited power.
   
To be more specific, 
   let $E(x_1,x_2, \ldots, x_n)$ be a multivariate objective function , or simply denoted as $E(x)$,
   where each variable $x_i$ has a finite domain $D_i$ of size $m_i$ ($m_i=|D_i|$).
We break the function into $n$ sub-objective functions $E_i$ ($i=1,2,\ldots,n$),
   such that $E_i$ contains at least variable $x_i$ for each $i$, 
   the minimization of each objective function $E_i$ (the sub-problem)
   is computational manageable in sizes and complexities, and
\begin{equation}
E (x) = \sum^{n}_{i = 1} E_i (x).
\label{decomposition}
\end{equation}

For example, a binary constraint optimization problem~(\ref{binary_cost_function}) has
a straight-forward decomposition:
\begin{equation}
E_i = C_i (x_i) + \sum_{j,~j \not = i} C_{ij} (x_i, x_j) \quad \mbox{for } i=1,2, \ldots, n\ . \nonumber
\label{decomposition_binary}
\end{equation}

The $n$ sub-problems can be described as:
\begin{equation}
\min_{x_j \in X_i} E_i, \quad \mbox{ for } i=1,2,\ldots,n \ , 
\label{sub-problem}
\end{equation}
where $X_i$ is the set of variables that sub-objective function $E_i$ contains.

Because of the interdependence of the sub-objective functions, 
   as in the case of the binary constraint-based function~(see Eq.~(\ref{binary_cost_function})),
   minimizing those sub-objective functions in such an independent way 
      can hardly yield a consensus in variable assignments.
For example, 
   the assignment for $x_i$ that minimizes $E_i$ can hardly be the same
      as the assignment for the same variable that minimizes $E_j$ if $E_j$ contains $x_i$.
We need to solve those sub-problems in a cooperative way 
   so that we can reach a consensus in variable assignments.

To do that, we can break the minimization of each sub-objective function~(see (\ref{sub-problem})) into two steps,
\[ \min_{x_i} \min_{x_j \in X_i \setminus x_i} E_i,  \quad \mbox{for } i=1,2, \ldots, n \ , \] 
where $X_i \setminus {x_i}$ denotes the set $X_i$ minuses $\{x_i\}$.

That is, first we optimize $E_i$ with respect to all variables that $E_i$ contains except $x_i$.
This gives us the intermediate solution in optimizing $E_i$, denoted as $c_i(x_i)$,
\begin{equation}
c_i (x_i) = \min_{x_j \in X_i \setminus{x_i}} E_i  \quad \mbox{for } i=1,2, \ldots, n \ .
\label{unary-constraint0}
\end{equation}
Second, we optimize $c_i(x_i)$ with respect to $x_i$,
\begin{equation}
\min_{x_i} c_i (x_i)\ .
\label{unary-constraint1}
\end{equation}

The intermediate solutions of the optimization, $c_i (x_i)$,
   is an unary constraint on $x_i$ introduced by the algorithm, 
   called the assignment constraint on variable $x_i$.
Given a value of $x_i$, $c_i (x_i)$ is the minimal value of $E_i$.
To minimize $E_i$, those values of $x_i$ which have smaller assignment constraint values $c_i (x_i)$
   are preferred more than those of higher ones.

To introduce cooperation in solving the sub-problems,
   we add the unary constraints $c_j(x_j)$, weighted by a real value $\lambda$, back to the right side of (\ref{unary-constraint0})
   and modify the functions~(\ref{unary-constraint0}) to be iterative ones:
\begin{equation}
c^{(k)}_i (x_i) = \min_{x_j \in  X_i \setminus{x_i}}\left( \left(1 - \lambda_k \right) E_i  +
\lambda_k \sum_{j} w_{ij} c^{(k-1)}_j(x_j)\right) \ , 
\label{cooperative_optimization}
\end{equation}
where $k$ is the iteration step, 
   $w_{ij}$ are non-negative weight values satisfying $\sum_i w_{ij} = 1$.
It has been found~\cite{Huang03Greece} that such a choice of $w_{ij}$ makes sure 
   the iterative update functions converge. 
The function at the right side of the equation is called the modified sub-objective function, 
   denoted as $\tilde{E}_i$.
   
By adding back $c_j(x_j)$ to $E_i$, 
   we ask the optimization of $E_i$ to compromise its solution 
   with the solutions of the other sub-problems.
As a consequence, the cooperation in the optimization of all the sub-objective functions ($E_i$s)
   is achieved.
This optimization process defined in (\ref{cooperative_optimization}) 
   is called the cooperative optimization of the sub-problems.
  
Parameter $\lambda_k$ in (\ref{cooperative_optimization}) controls the level of the cooperation at step $k$
and is called the cooperation strength, satisfying $0 \le \lambda_k < 1.$   
A higher value for $\lambda_k$ in (\ref{cooperative_optimization}) will weigh 
   the solutions of the other sub-problems $c_j(x_j)$
   more than the one of the current sub-problem $E_i$.
In other words, the solution of each sub-problem
   will compromise more with the solutions of other sub-problems.
As a consequence, 
   a higher level of cooperation in the optimization is reached in this case.

The update functions~(\ref{cooperative_optimization}) are a set of difference equations of 
   the assignment constraints $c_i(x_i)$.
Unlike conventional difference equations used by probabilistic relaxation algorithms~\cite{Rosenfeld76},
   cooperative computations~\cite{Marr76}, and Hopfield Networks~\cite{Hopfield82},
   this set of difference equations always has one and only one equilibrium given $\lambda$ and $w_{ij}$.
The computation converges to the equilibrium with an exponential rate, $\lambda$, regardless of initial conditions of $c^{(0)}_i(x_i)$.
Those computational properties will be shown in theorems in the next section
   and their proofs are provided in \cite{HuangBookCCO}.

By minimizing the linear combination of $E_i$ and $c_j(x_j)$, which are the intermediate solutions 
   for other sub-problems,
   we can reasonably expect that a consensus in variable assignments can be reached.
When the cooperation is strong enough, i.e., $\lambda_k \rightarrow 1$,
   the difference equations~(\ref{cooperative_optimization}) are dominated by the assignment constraints $c_j(x_j)$,
   it appears to us that the only choice for $x_j$ 
   is the one that minimizes $c_j(x_j)$
   for any $E_i$ that contains $x_j$.
That is a consensus in variable assignments.

Theory only guarantees the convergence of the computation 
   to the unique equilibrium of the difference equations.
If it converges to a consensus equilibrium, the solution, 
   which is consisted of the consensus assignments for variables,
   must be the global optimum of the objective function $E(x)$, guaranteed by theory (detail in the next section).
However, theory doesn't guarantee the equilibrium to be a consensus, 
   even by increasing the cooperation strength $\lambda$.
Otherwise, NP=P.
   

In addition to the cooperation scheme for reaching a consensus in variable assignments,
   we introduce another important operation of the algorithm, 
   called variable value discarding, at each iteration.
A certain value for a variable, say $x_i$, can be discarded
   if it has a assignment constraint value, $c_i(x_i)$ 
   that is higher than a certain threshold, $c_i(x_i) > t_i$,
   because they are less preferable in minimizing $E_i$ as explained before.
There do exist thresholds from theory for doing that (detail in the next section).
Those discarded values are those that can not be in any global optimal solution.
By discarding values, we can trim the search space. 
If only one value is left for each variable after a certain number of iterations
   using the thresholds provided theory, 
   they constitute the global optimal solution, guaranteed by theory~\cite{Huang03Greece}.
However, theory does not guarantee that one value is left for each variable in all cases.
Otherwise, NP=P.
This value discarding operation can be interpreted as neuron inhibition 
   following the winner-take-all principle
   if we implement this algorithm using neural networks.

By discarding values,
   we increase the chance of reaching a consensus equilibrium for the computation.
In practice, we progressively tighten the thresholds to discard more and more values 
   as the iteration proceeds to increase the chance of reaching a consensus equilibrium.
In the end, we leave only one value for each variable. 
Then, the final solution is a consensus equilibrium.

However, by doing that, such a final solution is not guaranteed to be the global optimum.
Nevertheless, in our experiments in solving large scale combinatorial optimization problems, 
   we found that the solution quality of this algorithm is still satisfactory,
   significantly better than that of other conventional optimization methods, such as simulated annealing and local search~\cite{Huang03Greece}.

\subsection{Definitions and Notations}

In the previous sub-section, we choose $w_{ij}$ 
   such that it is non-zero if $x_j$ is contained by $E_i$.
For a binary constraint optimization problem using the decomposition~(\ref{decomposition_binary}), 
   it implies that we choose $w_{ij}$ be non-zero if and only if $x_j$ is a neighbor of $x_i$.
However, theory tells us that this is too restrictive.
To make the algorithm to work,
   we only need to choose 
   $(w_{ij})_{n \times n}$ to be a propagation matrix defined as follows:

\begin{definition}
A propagation matrix $W=(w_{ij})_{n \times n}$
   is a irreducible, nonnegative, real-valued square matrix and satisfies
\[ \sum^n_{i=1} w_{ij}=1, \quad \mbox{ for } 1 \le j \le n\ . \]
\label{definition_propagation_matrix}
\end{definition}

A matrix $W$ is called reducible if there exists
   a permutation matrix $P$ such that $PWP^T$
   has the block form
\[\left(
    \begin{array}{cc}
      A & B \\
      O & C 
    \end{array}
\right)\ . \]

\begin{definition}
The system is called reaching a consensus solution
   if, for any $i$ and $j$ where $E_j$ contains $x_i$, 
   \[ \arg \min_{x_i} \tilde{E}_i = \arg \min_{x_i} \tilde{E}_j \ , \]
where $\tilde{E}_i$ is defined as the function to be minimized at the right side of Eq.~(\ref{cooperative_optimization}).
\label{consensus}
\end{definition}

\begin{definition}
An equilibrium of the system is  
	a solution to $c_i(x_i)$, $i = 1, 2, \ldots, n$, 
	that satisfies the difference equations~(\ref{cooperative_optimization}).
\end{definition}

To simplify the notations in the following discussions,
   let
\[ c^{(k)} = (c^{(k)}_1, c^{(k)}_2, \ldots,c^{(k)}_n). \]
Let $\tilde{x}^{(k)}_i = \arg \min_{x_i} c^{(k)}_i(x_i)$, 
   the favorable value for assigning variable $x_i$.
Let $\tilde{x}^{(k)} = (\tilde{x}^{(k)}_1, \tilde{x}^{(k)}_2, \ldots, \tilde{x}^{(k)}_n)$,
   a candidate solution at iteration $k$.
   
\section{Theoretical Foundations}

\subsection{General Properties}
The following theorem shows that 
   $c^{(k)}_i(x_i)$ for $x_i \in D_i$
   have a direct relationship to the lower bound on the objective function $E(x)$.

\begin{theorem} 
Given any propagation matrix $W$
   and the general initial condition $c^{(0)}=0$ or $\lambda_{1}=0$,
   $\sum_i c^{(k)}_i(x_i)$ is a lower bound function on $E(x_1,\ldots, x_n)$,
   denoted as $E^{(k)}_{-}(x_1,\ldots, x_n)$.
That is 
\begin{equation}
\sum_i c^{(k)}_i(x_i) \le E(x_1,x_2,\ldots, x_n),~~~~~ \mbox{for any $k\ge 1$}\ .
\label{up_bound}
\end{equation}
In particular, let $E^{*(k)}_{-}=\sum c^{(k)}_i(\tilde{x_i})$, 
   then $E^{*(k)}_{-}$ is a lower bound on the optimal cost $E^{*}$, 
   that is $E^{*(k)}_{-} \le E^{*}$.
\label{theorem_1}
\end{theorem}

Here, subscript ``-'' in $E^{*(k)}_{-}$ indicates that
   it is a lower bound on $E^{*}$.

This theorem tells us that $\sum c^{(k)}_i(\tilde{x_i})$ provides
   a lower bound on the objective function $E$.
We will show in the next theorem that
   this lower bound is guaranteed to be improved 
   as the iteration proceeds.
\begin{theorem}
Given any propagation matrix $W$,
   a constant cooperation strength $\lambda$,
   and the general condition $c^{(0)}=0$, 
   $\{E^{*(k)}_{-}|k \ge 0\}$
   is a non-decreasing sequence with upper bound $E^{*}$.
\label{theorem_3}
\end{theorem}

If a consensus solution is found at some step or steps,
   then we can find out the closeness between 
   the consensus solution and the global optimum in cost.
If the algorithm converges to a consensus solution, 
   then it must be the global optimum also.
The following theorem makes these points clearer.

\begin{theorem}
Given any propagation matrix $W$,
   and the general initial condition $c^{(0)}=0$ or $\lambda_1=0$.
If a consensus solution $\tilde{x}$ is found
   at iteration step $k_1$ and remains the same from step $k_1$ to step $k_2$,
then the closeness between the cost of $\tilde{x}$,
    $E(\tilde{x})$, and the optimal cost, $E^{*}$, satisfies 
   the following two inequalities,
\begin{equation}
0 \le E(\tilde{x})- E^{*} \le 
   \left(\prod^{k_2}_{k=k_1} \lambda_k\right) \left(E(\tilde{x})-E^{*(k_1-1)}_{-}\right),
\end{equation}
\begin{equation}
0 \le E(\tilde{x})- E^{*} \le 
   \frac{\prod^{k_2}_{k=k_1} \lambda_k}{1-\prod^{k_2}_{k=k_1}\lambda_k} (E^{*}-E^{*(k_1-1)}_{-})\ ,
\end{equation}
where $(E^{*}-E^{*(k_1-1)}_{-})$ is the difference 
   between the optimal cost $E^{*}$ and
   the lower bound on the optimal cost, $E^{*(k_1-1)}_{-}$,
   obtained at step $k_1 - 1$.
When $k_2-k_1 \rightarrow \infty$ and $1 - \lambda_k \ge \epsilon > 0$ 
   for $k_1 \le k \le k_2$, $E(\tilde{x}) \rightarrow E^{*}$.
\label{theorem_2}
\end{theorem}  

\subsection{Convergence Properties}

The performance of the cooperative algorithm further depends on
   the dynamic behavior of the difference equations~(\ref{cooperative_optimization}).
Its convergence property
   is revealed in the following two theorems.
The first one shows that, 
   given any propagation matrix 
   and a constant cooperation strength,
   there does exist a solution
   to satisfy the difference equations (\ref{cooperative_optimization}).
The second part shows that the cooperative algorithm 
   converges exponentially to that solution.

\begin{theorem}
Given any symmetric propagation matrix $W$
   and a constant cooperation strength $\lambda$,
   then Difference Equations~(\ref{cooperative_optimization})
   have one and only one solution, denoted as $(c^{(\infty)}_i(x_i))$
   or simply $\mbox{\boldmath c}^{(\infty)}$.
\label{theorem_7}
\end{theorem}

\begin{theorem}
Given any symmetric propagation matrix $W$ and 
   a constant cooperation strength $\lambda$,
   the cooperative algorithm,
   with any choice of the initial condition $c^{(0)}$,
   converges to $c^{(\infty)}$ with an exponential convergence rate $\lambda$.
That is
\begin{equation}
\|c^{(k)}-c^{(\infty)}\|_{\infty} \le 
   \lambda^k \|c^{(0)}-c^{(\infty)}\|_{\infty}\ .
\end{equation}
\label{theorem_8}
\end{theorem}

This theorem is called the convergence theorem.
It indicates that
   our cooperative algorithm is stable and
   has a unique attractor, $c^{(\infty)}$.
Hence, the evolution of our cooperative algorithm is robust,
   insensitive to perturbations, and
   the final solution of the algorithm
   is independent of initial conditions.
In contrast, conventional algorithms
   based on iterative improvement
   have many local attractors due to the local minima problem.
The evolutions of these algorithms are sensitive 
   to perturbations,
   and the final solutions of these algorithms
   are dependent on initial conditions.
   
\subsection{Necessary Conditions}
 The two necessary conditions provides in this subsection
   allows us to discard variable values that can not be in any global optimum.
 
\begin{theorem} 
Given a propagation matrix $W$,
   and the general initial condition $c^{(0)}=0$ or $\lambda_{1}=0$.
If value $x^{*}_i$ ($x^{*}_i \in D_i$)
   is in the global optimum, 
   then $c^{(k)}_i(x^{*}_i)$, for any $k \ge 1$,
   must satisfy the following inequality,
\begin{equation}
c^{(k)}_i(x^{*}_i)\le (E^{*}-E^{*(k)}_{-})+c^{(k)}_i(\tilde{x}^{(k)}_i)
\label{eqn_necessary_condition_1}
\end{equation}
where $E^{*(k)}_{-}$ is, as defined before, 
   a lower bound on $E^{*}$ 
   obtained by the cooperative system at step $k$.
\label{necessary_condition_1}
\end{theorem}

\begin{theorem}
Given a symmetric propagation matrix $W$ and
   the general initial condition $c^{(0)}=0$ or $\lambda_1=0$.
If value $x^{*}_i$ ($x^{*}_i \in D_i$) 
   is in the global optimum,
then $c^{(k)}_i(x^{*}_i)$ must satisfy the following inequality,
\begin{equation}
\label{eqn_necessary_condition_2}
c^{(k)}_i(x^{*}_i) \le \frac{E^{*}}{n}+\sqrt{\frac{n-1}{n}} |\alpha^{(k)}_2| E^{*}
\end{equation}
Here $\alpha^{(k)}_2$ is computed by the following recursive function:
\begin{displaymath}
\left\{ \begin{array}{l}
   \alpha^{(1)}_2 = \lambda_1 \alpha_2 + (1 - \lambda_1) \\
   \alpha^{(k)}_2 = \lambda_k \alpha_2 \alpha^{(k-1)}_2 +  (1 - \lambda_{k})
       \end{array}
\right.
\end{displaymath}
where $\alpha_2$ is the second largest eigenvalue of the propagation matrix $W$.

For the particular choice of W=$\frac{1}{n}(1)_{n \times n}$, 
\[ \alpha^{(k)}_2 = (1 - \lambda_{k}) \]
and 
\begin{equation}
c^{(k)}_i(x^{*}_i) \le \frac{E^{*}}{n}+\sqrt{\frac{n-1}{n}} (1 - \lambda_{k}) E^{*}.
\end{equation}
\label{necessary_condition_2}
\end{theorem}

Inequality~(\ref{eqn_necessary_condition_1}) and Inequality~(\ref{eqn_necessary_condition_2})
   provide two criteria
   for checking if a value can be in some global optimum.
If either of them is not satisfied, 
   the value can be discarded from the value set to reduce the search space.

Both thresholds in (\ref{eqn_necessary_condition_1}) and (\ref{eqn_necessary_condition_2}) 
   become tighter and tighter as the iteration proceeds.
Therefore, more and more values can be discarded 
   and the search space can be reduced.
With the choice of the general initial condition $c^{(0)}=0$,
   the right hand side of (\ref{eqn_necessary_condition_1}) decreases
   as the iteration proceeds because of the property of $E^{*(k)}_{-}$ revealed by Theorem~\ref{theorem_3}.
With the choice of a constant cooperation strength $\lambda$,
   and suppose $W \not = \frac{1}{n}(1)_{n \times n}$, 
   then $\alpha_2 > 0$ and
   $\{\alpha^{(k)}_2|k \ge 1\}$ is a monotonic decreasing sequence satisfying
\begin{equation}
\frac{1 - \lambda}{1-\lambda \alpha_2} <\alpha^{(k)}_2 \le (1 - \lambda) + \lambda \alpha_2 
\end{equation}
This implies that the right hand side of (\ref{eqn_necessary_condition_2})
   monotonically decreases as the iteration proceeds.
   
\section{Case Studies in Computer Vision}

The proposed algorithm 
   has outperformed many well-known optimization algorithms in solving real optimization problems
 	in computer vision\cite{Huang03Greece,Huang03Turkey}, image processing\cite{Huang04ICIP}, and data communications.
These experiment results give strong evidence of the algorithm's considerable potential.

We provides in this section the performance comparison 
	of the new Hopfield networks with cooperative optimization
	and Boltzmann machine network 
	for stereo matching~\cite{Scharstein2002,Boykov2001,Kolmogorov04,Szeliski:va99}.
The Boltzmann machine is simply a discrete time Hopfield network in which the dynamic function of each neuron
	is defined by simulated annealing~\cite{Kirkpatrick83}.
Simulated annealing is a well-known optimization method which is based on stochastic local optimization.

\begin{figure}
\center{{\epsfxsize 2.7cm \epsffile{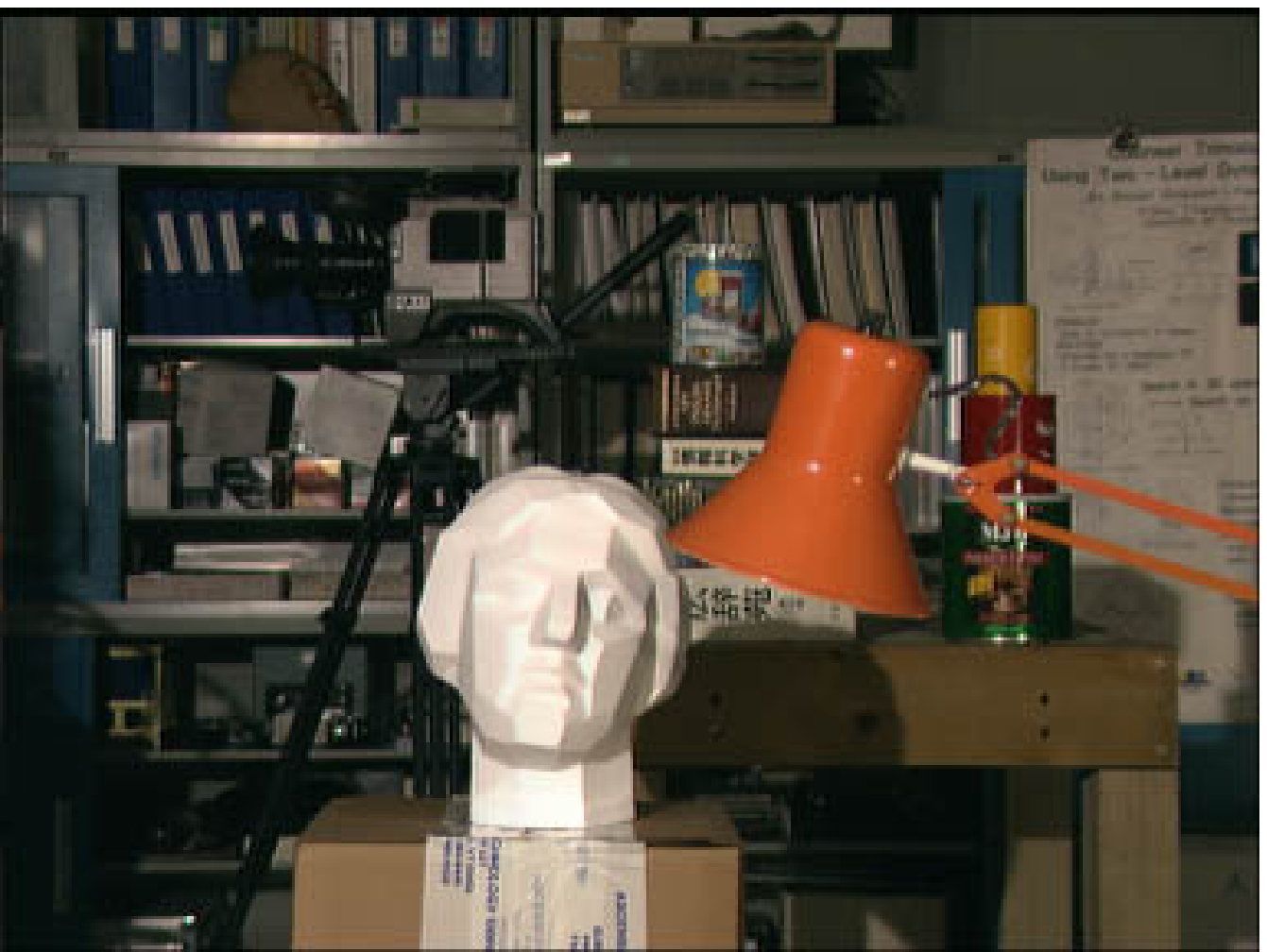}} {\epsfxsize 2.7cm \epsffile{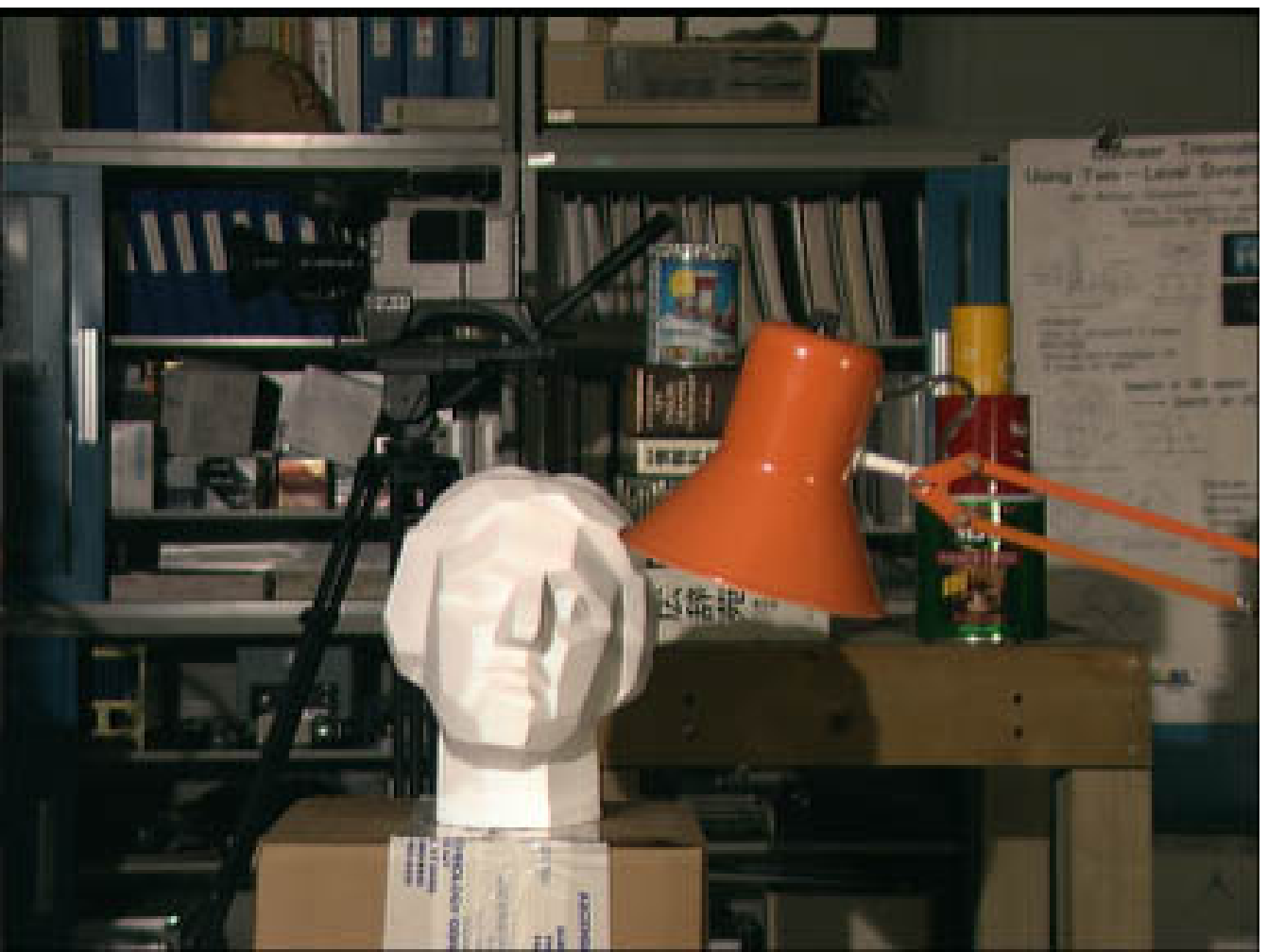}} }
\caption{A pair of images for stereo matching.}
\label{twoimages}
\end{figure}

Stereo vision is an important process in the human visual perception.
As of now, there is still a lack of satisfactory computational neural model for it.
To understand such an important process, 
	people treat stereo vision as stereo matching.
Stereo matching is to use a pair of 2-D images of the same scene taken at the same time but two different locations 
	to recover the depth information 
	of the scene (see Fig.~\ref{twoimages}).

\begin{figure}
\center{{\epsfxsize 2.7cm \epsffile{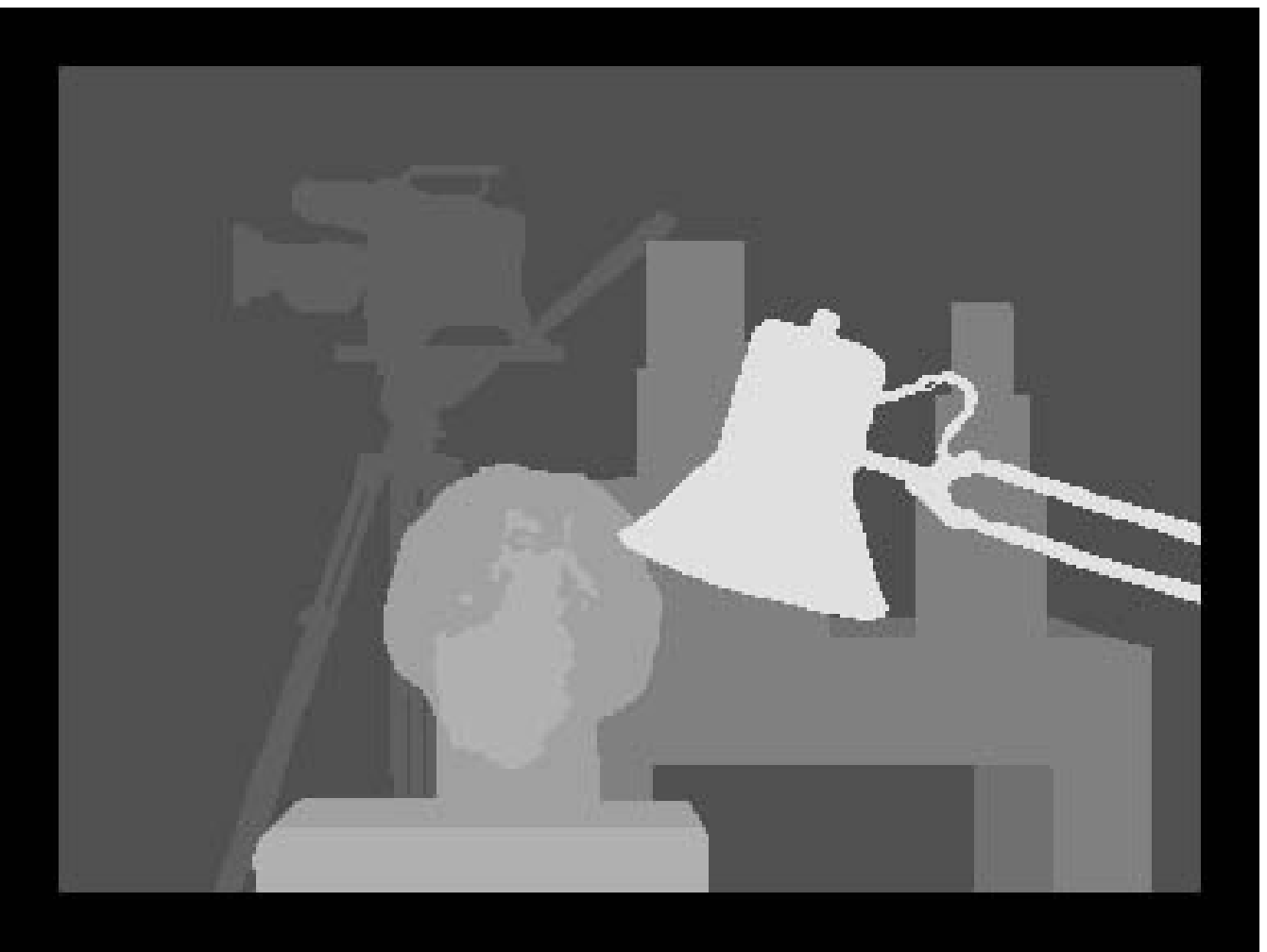}} {\epsfxsize 2.7cm \epsffile{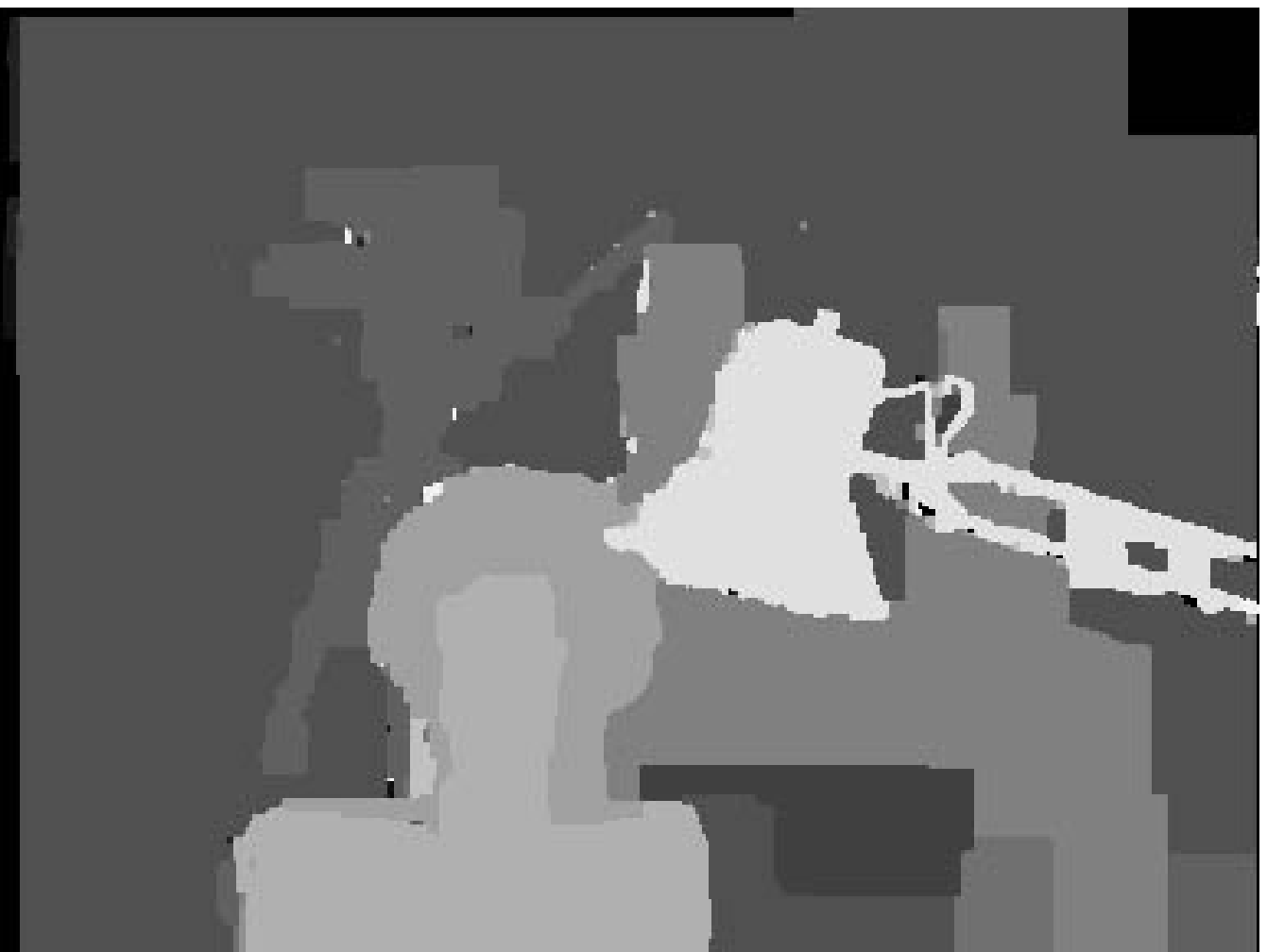}} {\epsfxsize 2.7cm \epsffile{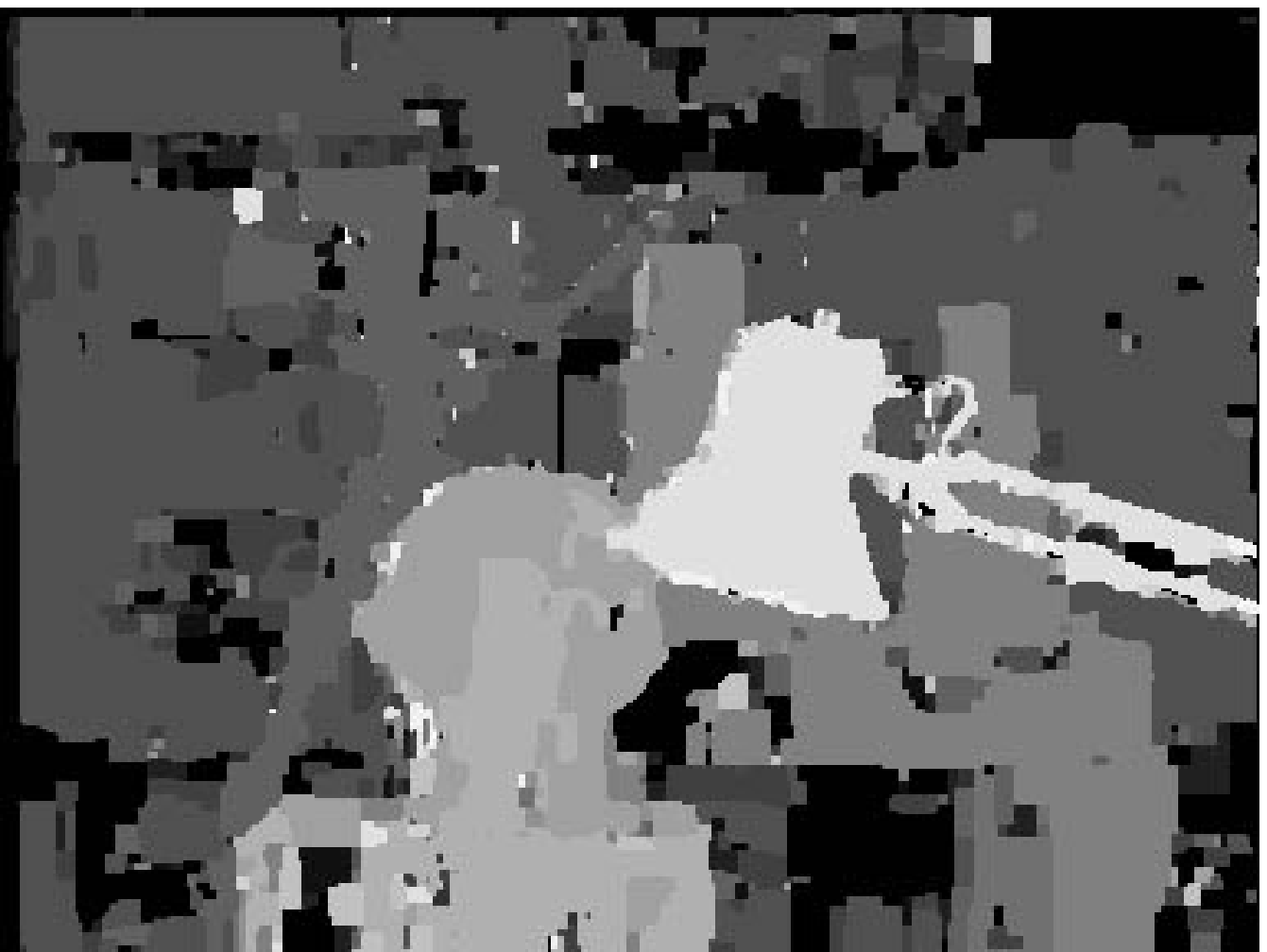}}}
\center{{\epsfxsize 2.7cm \epsffile{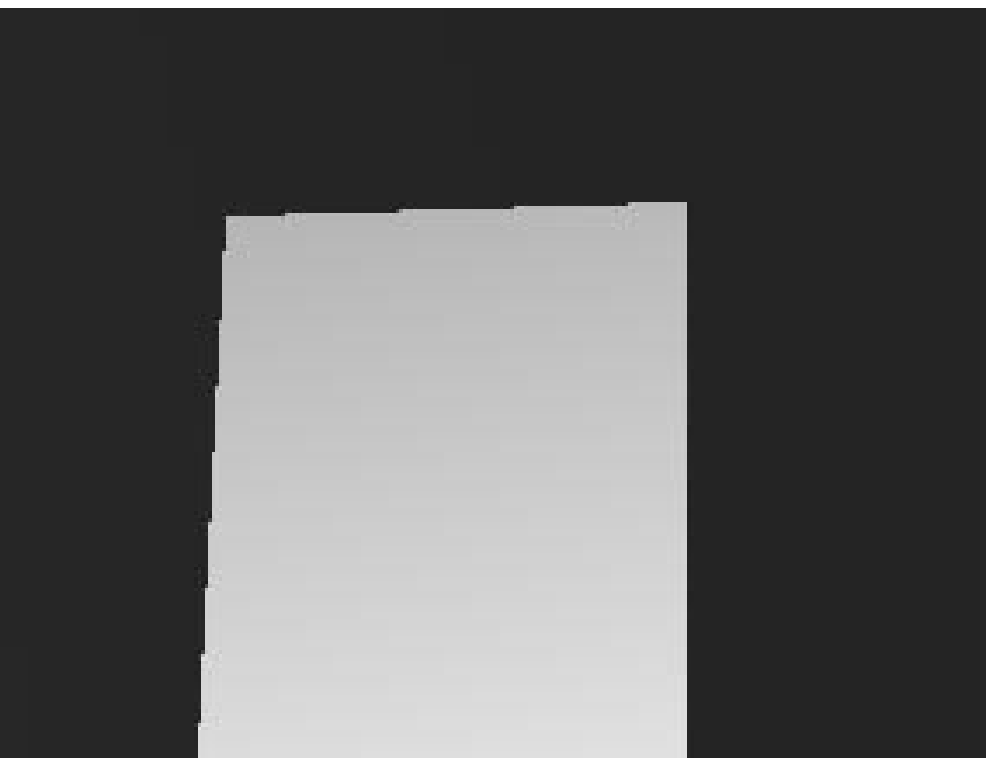}} {\epsfxsize 2.7cm \epsffile{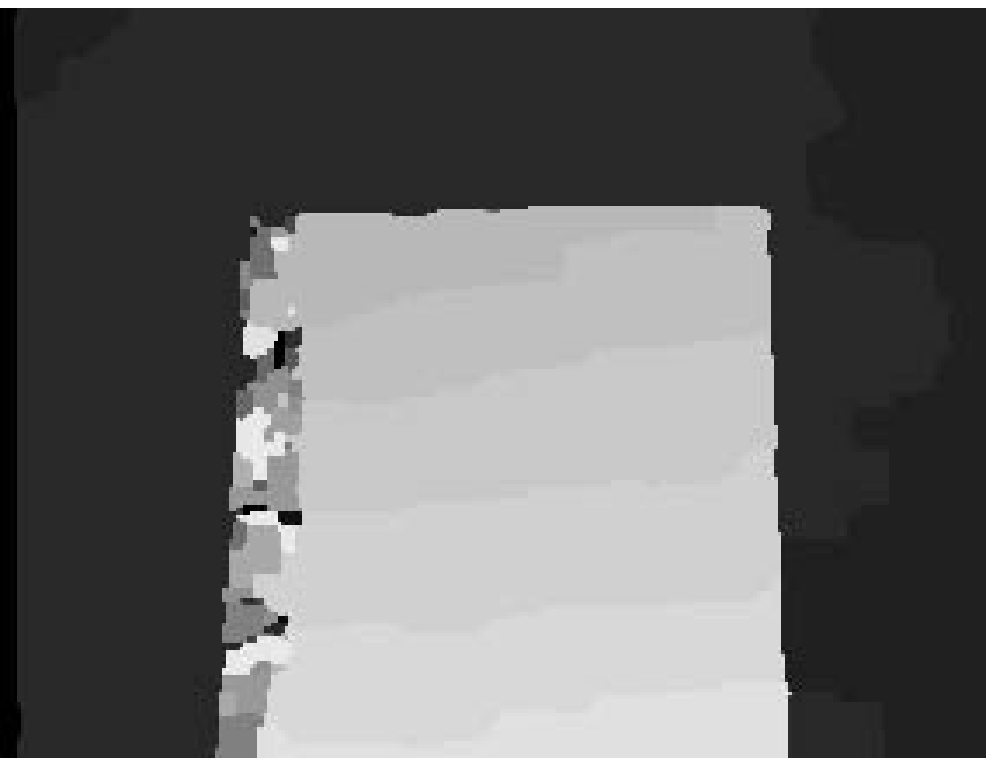}} {\epsfxsize 2.7cm \epsffile{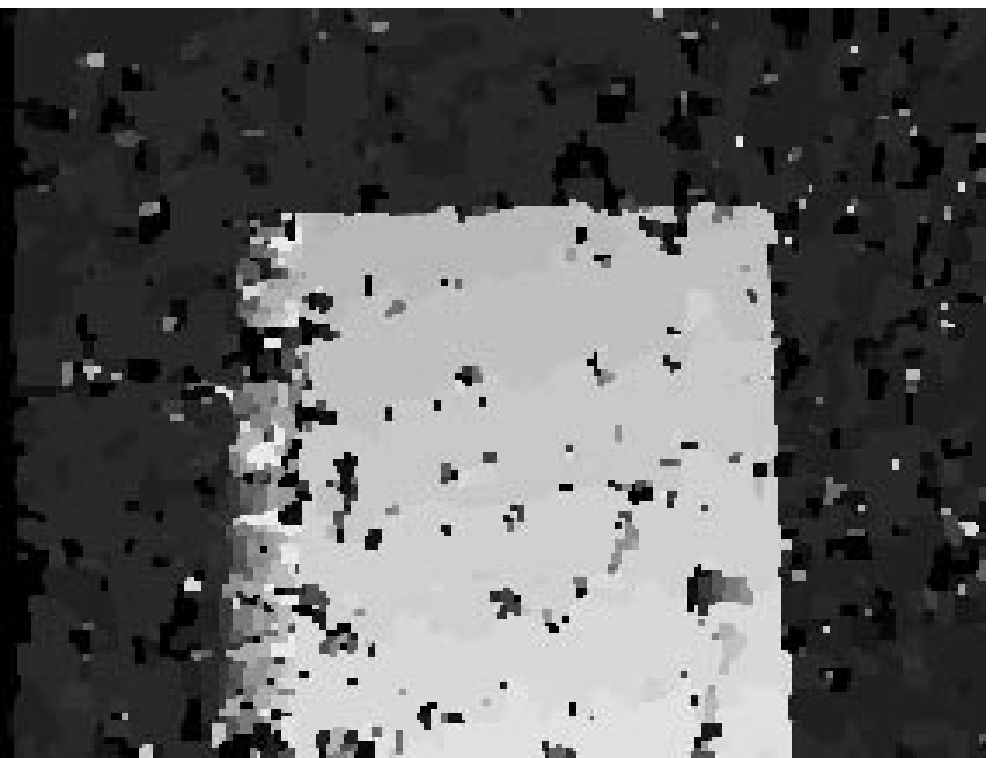}}}
\center{{\epsfxsize 2.7cm \epsffile{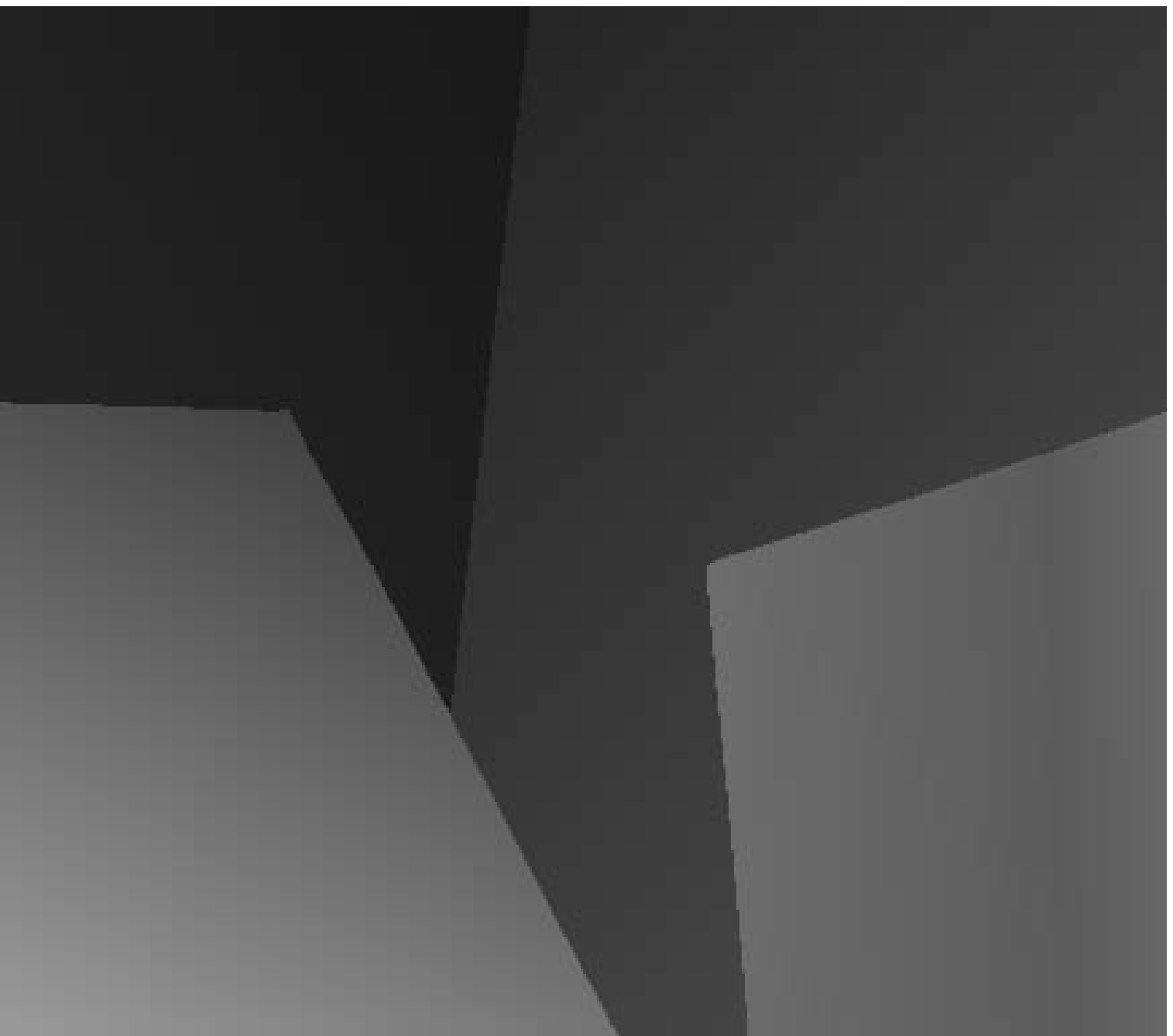}} {\epsfxsize 2.7cm \epsffile{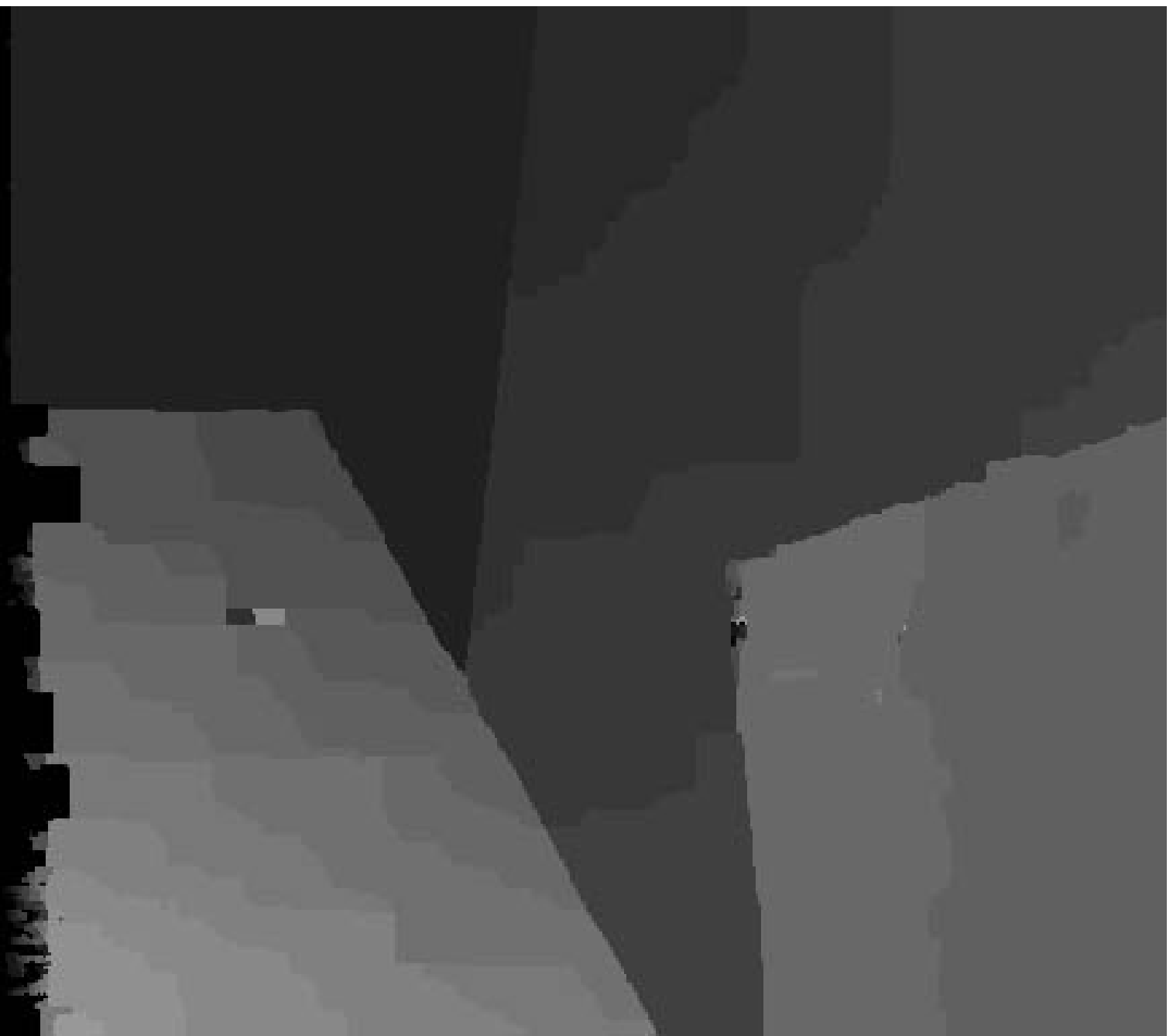}} {\epsfxsize 2.7cm \epsffile{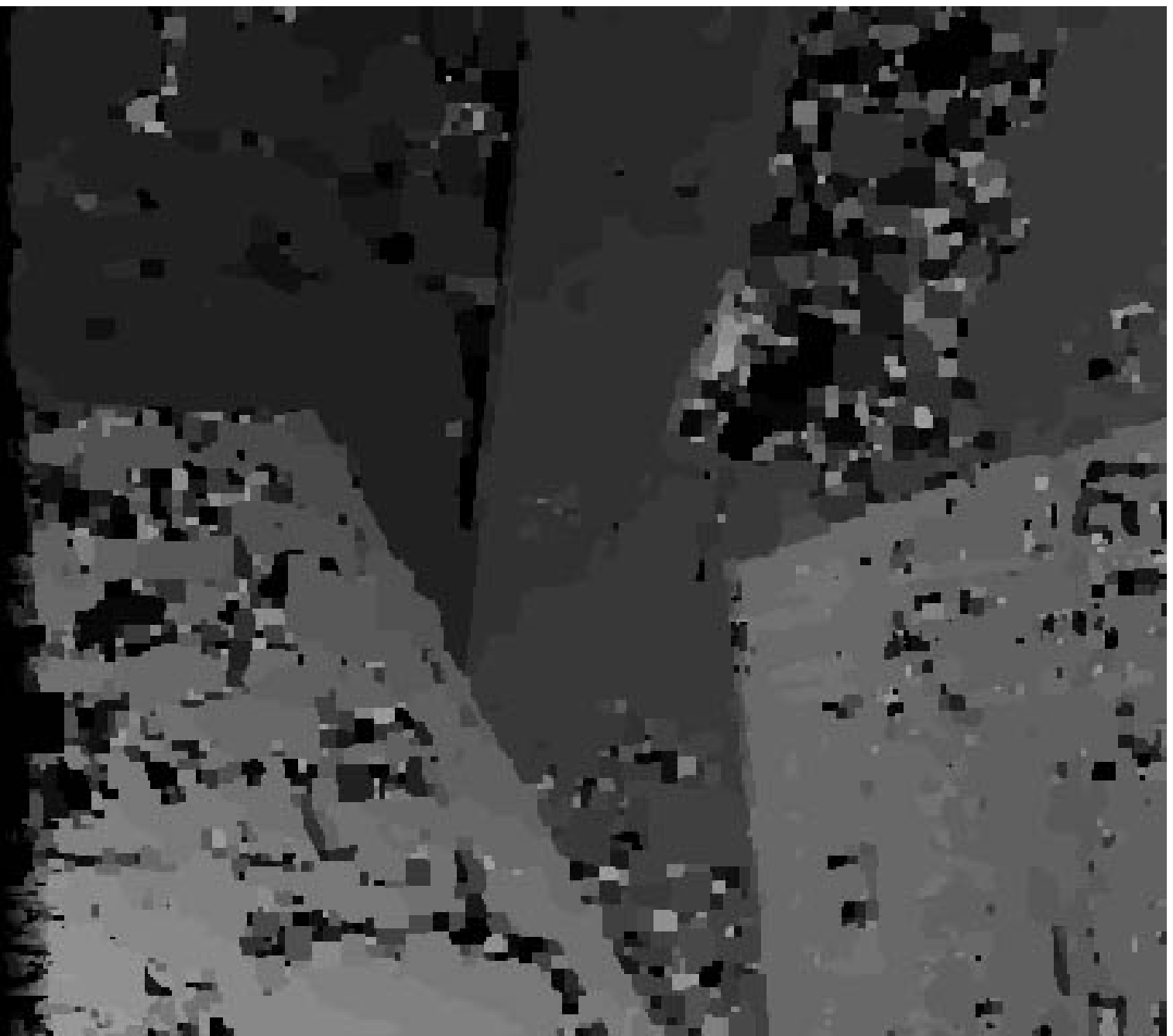}}}
\center{{\epsfxsize 2.7cm \epsffile{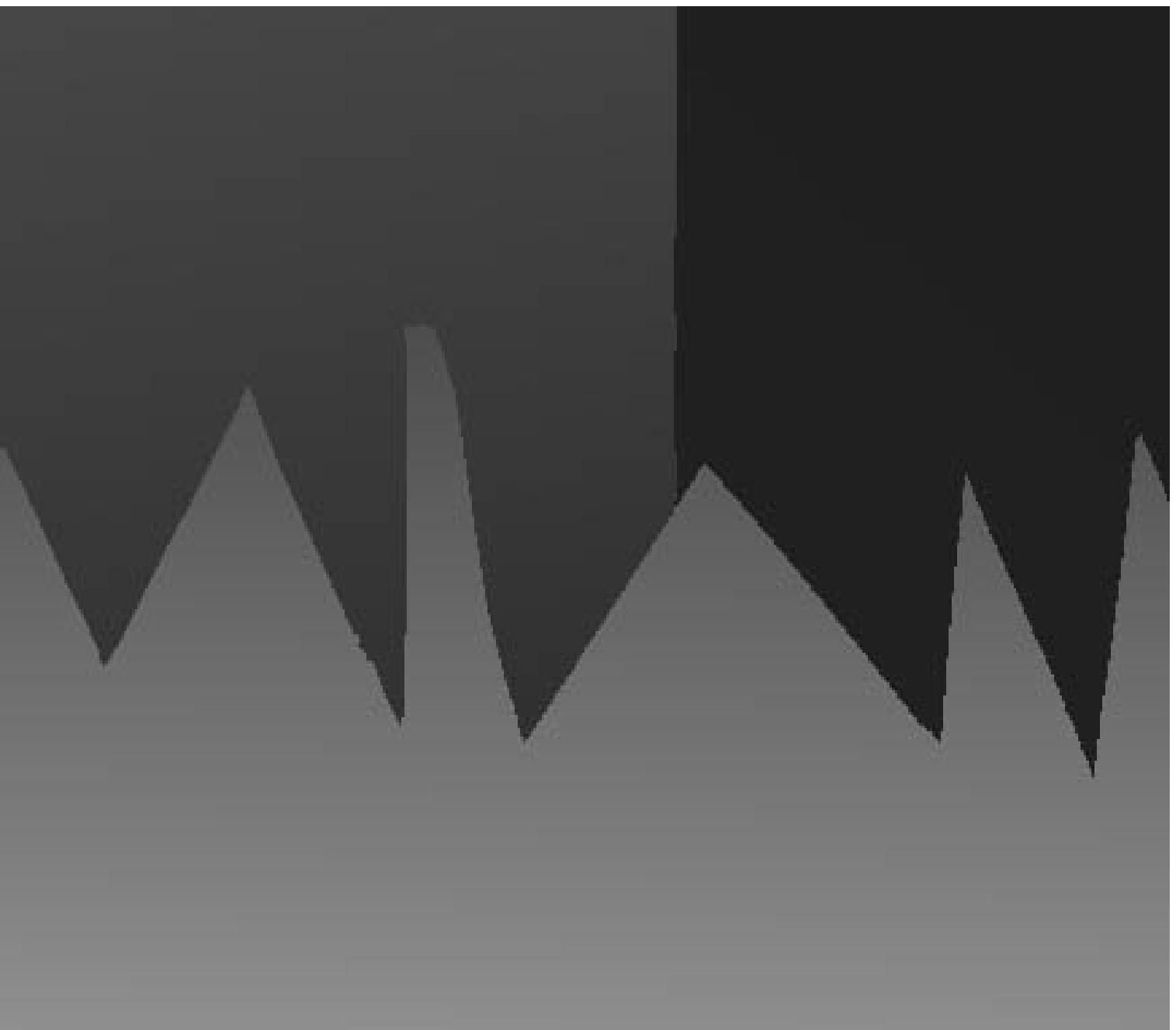}} {\epsfxsize 2.7cm \epsffile{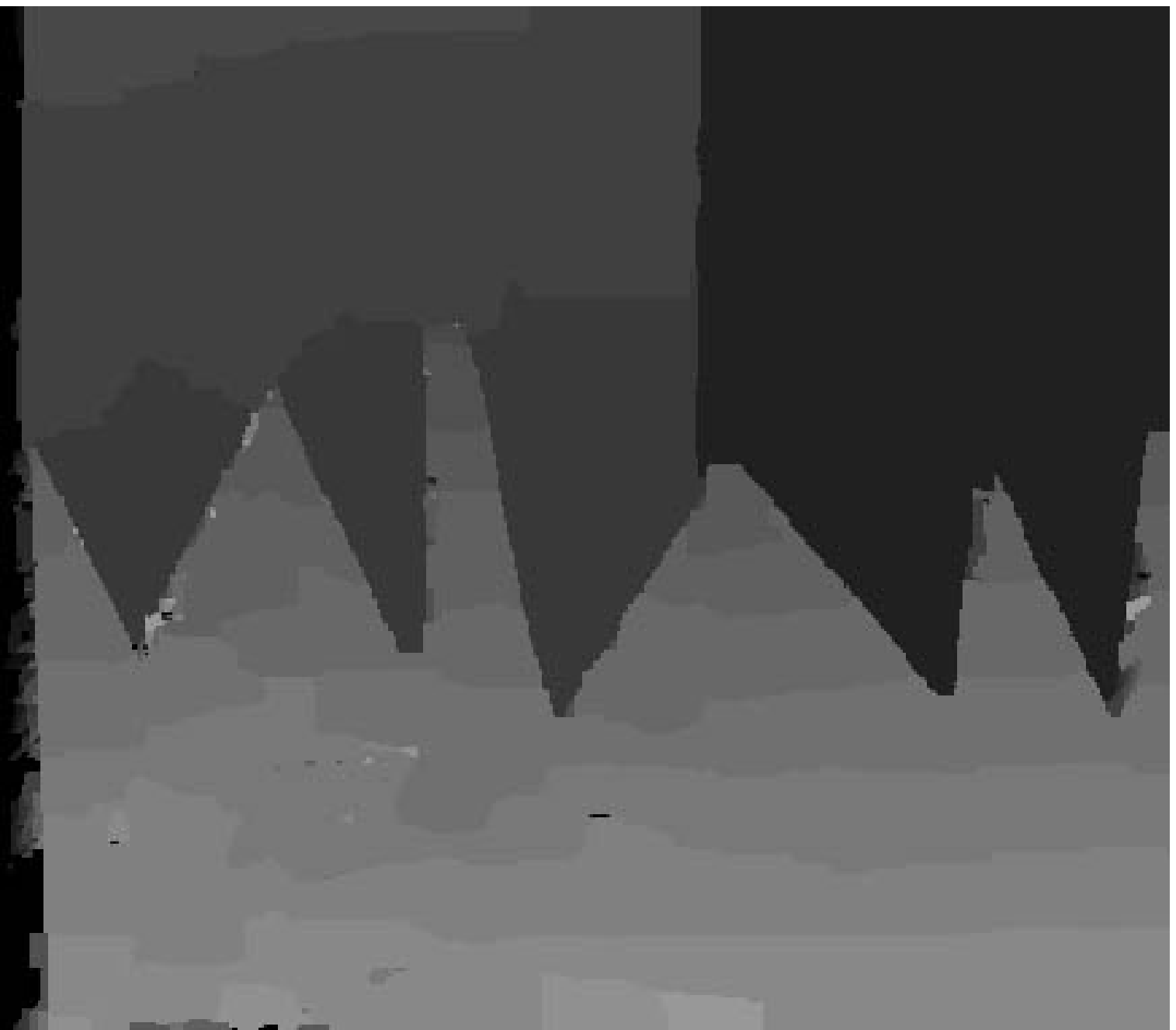}} {\epsfxsize 2.7cm \epsffile{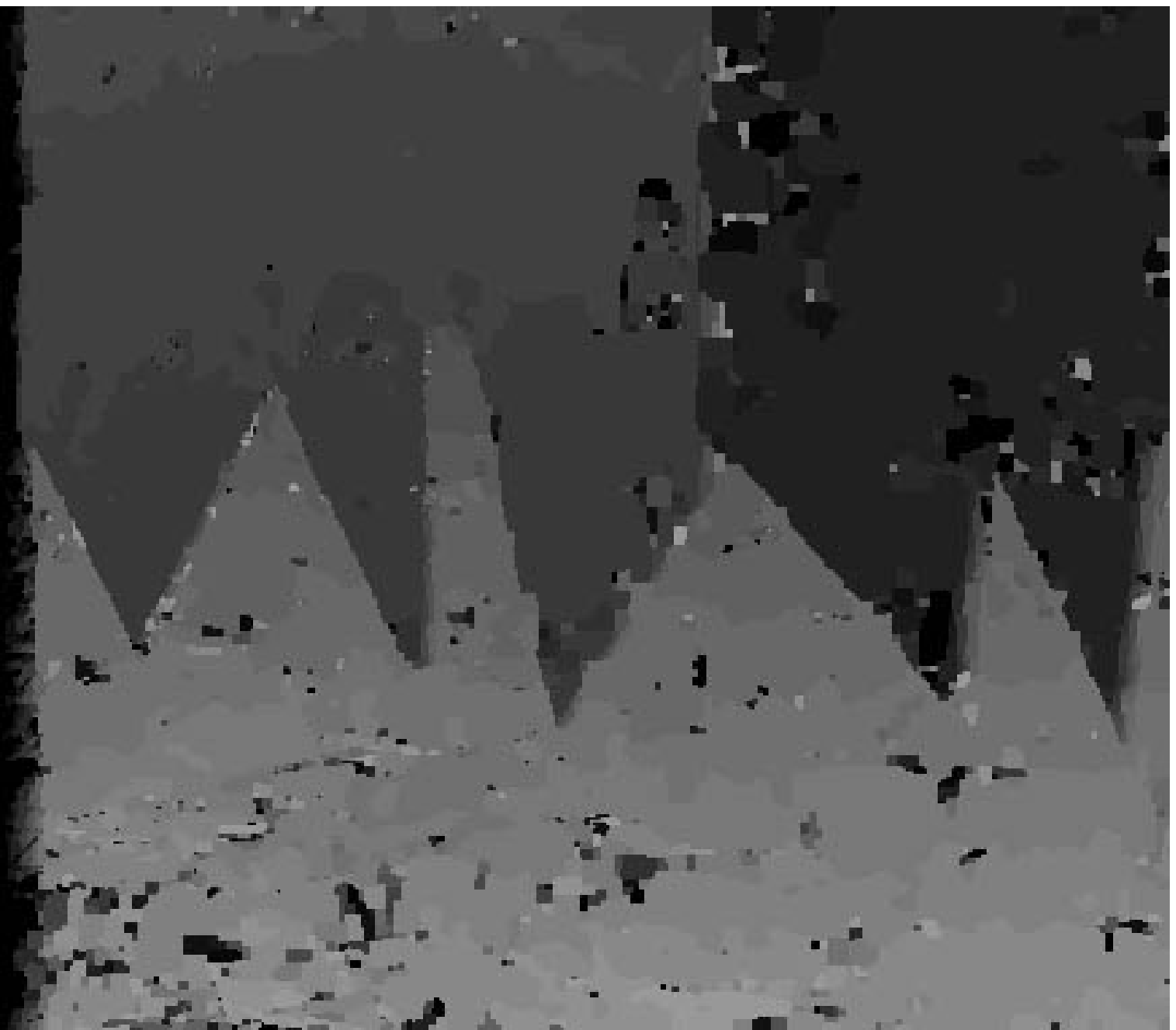}}}
\caption{The performances of neural networks with different dynamics for four real instances of stereo matching. 
The ground truth (left). 
The results of the Hopfield nework with cooperative optimization (middle). 
The results of the Boltzmann machine network (right).}
\label{groundtruth}
\end{figure}

Instead of using toy problems, 
	we tested both types of neural networks with real problems.
Four pairs of images including the one shown in Fig.~\ref{twoimages} are used in our experiments.
The ground truth, the depth images obtained by Boltzmann machine 
	and by the new Hopfield network with cooperative optimization are shown in Fig.~\ref{groundtruth}.
Clearly, the results of new Hopfield with cooperative optimization 
	are much cleaner, much smoother, and much better than the results of Boltzmann machine.

Table~\ref{table_1} lists the minimal energies found by the two types of neural networks.
Those found the new Hopfield network are much lower those found by Boltzmann machine.

\begin{table}
\begin{center}
\begin{tabular}{ccc}  
\hline\noalign{\smallskip}
& \multicolumn{2}{c}{Minimal Energies ($\times 10^3$)} \\
Image & Boltzmann Machine & New Hopfield Network\\
\noalign{\smallskip}
\hline
\noalign{\smallskip}
Map & 580 & 329 \\
Sawtooth & 182 & 143 \\
Tsukuba & 781 & 518 \\
Venus & 197 & 125\\ 
\hline
\end{tabular}
\end{center}
\caption{The minimal energies found by neural networks with different dynamics. }
\label{table_1}
\end{table}

The iteration time is $16$ for the new Hopfield network and $100$ for Boltzmann network.
In each iteration, all neurons are updated once.
On average, the new Hopfield network is three times faster than Boltzmann machine in our simulation.
Another big advantage of the new Hopfield network over Boltzmann machine is its inherited parralism.
In each iteration, all neurons in the new Hopfield network can be updated fully in parallel.
This feature, together with the excellent performance of the new Hopfield network 
	offer us commerial pontential 
	in implementing stereo vision capability for robots and unmanned vehicles.
	
\section{Conclusions}

This paper presented a neural network implementation of a new powerful cooperative algorithm 
   for solving combinatorial optimization problems.
It fixes many problems of the Hopfield network in theoretical, performance, and implementation perspectives.
Its operations are based on parallel, local iterative interactions.
The proposed algorithm has many important computational properties absent in existing optimization methods.
Given an optimization problem instance, the computation always has a unique equilibrium 
   and converges to it with an exponential rate regardless of initial conditions and perturbations.
There are sufficient conditions~\cite{Huang03Greece} for identifying global optimum and
   necessary conditions for trimming search spaces.
In solving large scale optimization problems in computer vision, 
   it significantly outperformed classical optimization methods, such as simulated annealing 
   and local search with multi-restarts.   

One of the key processes of cooperative computation is value discarding.
This is the same in principle as the inhibition process
   used by Marr and Poggio in \cite{Marr76}, and Lawrence and Kanade in \cite{Lawrence2000}.
The inhibition process 
   makes the cooperative computation fundamentally different 
   from the most known optimization methods.
As Steven Pinker pointed out in his book ``How the Mind Works'',
   the cooperative optimization captures the flavor of the brain's computation of stereo vision.
It has many important computational properties 
   not possessed by conventional ones.
They could help us in understanding cooperative computation possibly used by human brains 
    in solving early vision problems.
 
\nocite{Haykin99}
\nocite{Huang03Greece,Huang03Turkey}
\nocite{Michalewicz02,CPapadimitriou98}
\nocite{Michalewicz02,CPapadimitriou98}
\nocite{Holland92}
\nocite{Lawler66,Coffman76}
\nocite{Coffman76} 
\nocite{Marr76,Rosenfeld76,Hopfield82}

\bibliographystyle{IEEEtran}

\end{document}